\documentclass[conference]{IEEEtran}
\IEEEoverridecommandlockouts
\usepackage{cite}
\usepackage{amsmath,amssymb,amsfonts}
\usepackage{algorithmic}
\usepackage{graphicx}
\usepackage{textcomp}
\usepackage{xcolor}
\def\BibTeX{{\rm B\kern-.05em{\sc i\kern-.025em b}\kern-.08em
    T\kern-.1667em\lower.7ex\hbox{E}\kern-.125emX}}
\begin{document}

\title{Company classification using zero-shot learning
}

\author{\IEEEauthorblockN{1\textsuperscript{st} Maryan Rizinski}
\IEEEauthorblockA{\textit{Faculty of Comp. Sci. and Eng.} \\
\textit{Ss. Cyril and Methodius University}\\
Skopje, North Macedonia \\
rizinski@bu.edu}
\and
\IEEEauthorblockN{2\textsuperscript{nd} Andrej Jankov}
\IEEEauthorblockA{\textit{Faculty of Comp. Sci. and Eng.} \\
\textit{Ss. Cyril and Methodius University}\\
Skopje, North Macedonia \\
andrej.jankov@students.finki.ukim.mk}
\and
\IEEEauthorblockN{3\textsuperscript{nd} Vignesh Sankaradas}
\IEEEauthorblockA{\textit{Department of Computer Science} \\
\textit{Boston University, Metropolitan College}\\
Boston, MA 02215, USA \\
vigneshs@bu.edu}
\and
\IEEEauthorblockN{4\textsuperscript{nd} Eugene Pinsky}
\IEEEauthorblockA{\textit{Department of Computer Science} \\
\textit{Boston University, Metropolitan College}\\
Boston, MA 02215, USA \\
epinsky@bu.edu}
\and
\IEEEauthorblockN{5\textsuperscript{rd} Igor Miskovski}
\IEEEauthorblockA{\textit{Faculty of Comp. Sci. and Eng.} \\
\textit{Ss. Cyril and Methodius University}\\
Skopje, North Macedonia \\
igor.miskovski@finki.ukim.mk}
\and
\IEEEauthorblockN{6\textsuperscript{th} Dimitar Trajanov}
\IEEEauthorblockA{\textit{Faculty of Comp. Sci. and Eng.} \\
\textit{Ss. Cyril and Methodius University}\\
Skopje, North Macedonia \\
dimitar.trajanov@finki.ukim.mk}
}

\maketitle

\begin{abstract}
In recent years, natural language processing (NLP) has become increasingly important in a variety of business applications, including sentiment analysis, text classification, and named entity recognition. In this paper, we propose an approach for company classification using NLP and zero-shot learning. Our method utilizes pre-trained transformer models to extract features from company descriptions, and then applies zero-shot learning to classify companies into relevant categories without the need for specific training data for each category. We evaluate our approach on a dataset obtained through the Wharton Research Data Services (WRDS), which comprises textual descriptions of publicly traded companies. We demonstrate that the approach can streamline the process of company classification, thereby reducing the time and resources required in traditional approaches such as the Global Industry Classification Standard (GICS). The results show that this method has potential for automation of company classification, making it a promising avenue for future research in this area.
\end{abstract}


\section{Introduction}
\label{sec:intro}

In financial research, company classification is a popular approach that involves grouping similar companies into categories or clusters \cite{bhojraj2003s, lyocsa2009industry}. This process of classifying companies into discrete categories has numerous practical applications for financial researchers, analysts, decision-makers, and investors. For example, it can help manage portfolio risk, facilitate relative valuation, and enable peer-group comparisons \cite{chan2007industry}. Additionally, it can aid in the analysis of the effects of corporate reorganizations, changes in financial and investment policies, and the evaluation of the performance of a specific company against a set of similar companies. Beyond the financial sector, company classification can be useful in generating prospective leads for sales and marketing teams, identifying new clients for insurance companies, and pinpointing competitors for corporations. Investment banks and venture capital firms can also benefit from company classification by understanding the distribution of companies among different industries \cite{wood2017automated}.

The task of company classification has traditionally relied on established standards, such as the Standard Industrial Classification (SIC), the North American Industry Classification System (NAICS), the Fama and French (FF) model, and the Global Industry Classification Standard (GICS), among others. Nonetheless, these methods have several limitations that hinder their effectiveness. One significant challenge is the lack of interoperability between different schemes, which results in classification inconsistencies due to vendor-specific assignments. Furthermore, the creation of existing standards is a laborious and time-consuming process that requires input from human experts, making them prone to subjectivity.

As products and services become increasingly complex, updating classification schemes becomes a challenging task. The dynamic market environment in which companies operate causes frequent changes in their business, affecting their industry affiliation. The existing classification standards are static and are unable to keep up with the fast-changing environments. The current schemes heavily rely on self-reporting and manual entry, resulting in slow, costly, and ineffective updates when adapting to the changed business landscape. In the absence of capabilities for real-time updates, these existing standards may not be the optimal choice in various application settings, thereby emphasizing the need to explore automation techniques.

The recent advancements in machine learning (ML) and natural language processing (NLP) can be explored to address the limitations of the traditional standards for company classification, and hold promises for reducing costs, complexity, and manual labor. In particular, text classification using NLP methods has achieved significant progress over the past decade. Large-scale pre-trained transformer models have revolutionized the field of text classification, enabling successful implementations across various application domains such as machine translation, text summarization, and sentiment analysis. These models witnessed successful deployments in systems that need scalability and real-time analysis, making them also valuable to address the problem of company classification.

\begin{table}[htbp!]
\caption{GICS taxonomy illustrating the names of the four classification levels, from broadest to narrowest, the number of categories for each classification level and the number of digits used to represent each level.}
\begin{center}
\begin{tabular}{cccc}
\hline
\multicolumn{4}{c}{\textbf{GICS taxonomy}}                                                                                      \\ \hline
Level               & Title          & \begin{tabular}[c]{@{}c@{}}Number of\\ categories\end{tabular} & Digits         \\ \hline
Level 1 (broadest)  & Sector         & 11                                                             & first 2 digits \\
Level 2             & Industry Group & 24                                                             & first 4 digits \\
Level 3             & Industry       & 64                                                             & first 6 digits \\ 
Level 4 (narrowest) & Sub-industry   & 139                                                            & all 8 digits   \\ \hline
\end{tabular}
\end{center}
\label{table:gics}
\end{table}

In this paper, we explore the use of zero-shot classification using the model valhalla/distilbart-mnli-12-3 on the Wharton Research Data Services (WRDS) dataset. The WRDS dataset contains names and textual descriptions of 34,338 companies that are classified as per the GICS index. In the classification experiment, we calculate standard metrics such as precision, recall, F1 score and support for each classification category as well as for the overall model. Our purpose is to evaluate the potential of zero-shot learning and use it as a basis for future research with unsupervised and supervised learning.

The organization of the paper is as follows. Section \ref{sec:background} focuses on background information and preliminaries. We review related work in the literature in Section \ref{sec:related_work}. In section \ref{sec:methodology}, we describe our NLP-based experiment and dataset for company classification using zero-shot learning and discuss the obtained results. Section \ref{sec:conclusion} concludes the paper.

\section{Standards for company classification}
\label{sec:background}

\subsection{Definition of company classification}

Company classification, also known as industry classification, involves categorizing companies based on their business activities, industry, and other relevant factors. The objective of this process is to group similar companies together and distinguish them from others based on several comparison parameters \cite{porter1980techniques}. Homogeneity is a crucial criterion for selecting an industry classification standard from the available options, and it is typically evaluated using various approaches. For instance, \cite{chan2007industry} suggests using stock return co-movement, while \cite{bhojraj2003s} recommends utilizing 12 fundamental variables. By segmenting the market into partitions with distinct business and financial characteristics \cite{phillips2016industry, yang2016automatic}, these classifications provide a framework for understanding the similarities and differences among companies. Ultimately, the goal is to identify groups of businesses that engage in similar market activities and have comparable market conditions.

\subsection{Mainstream standards}

Industry classification standards are an essential tool for economic analysis, financial research, and policy-making. Among the most prominent industry classification systems are the Standard Industrial Classification (SIC), the North American Industry Classification System (NAICS), Fama French (FF), and the Global Industry Classification Standard (GICS). The SIC system, established in the 1930s, is the oldest of the four and underwent periodic revisions to adapt to changes in the economy. However, these efforts were insufficient, prompting the governmental statistical agencies of the United States, Canada and Mexico, to collaborate on a joint initiative to improve the SIC system and create a more comprehensive and unified classification scheme across North America. The result was the creation of the NAICS system in 1999. The 2017 edition of the NAICS taxonomy partitions the North American economy into 1057 industries with a six-digit code for each industry. The NAICS system includes 20 sectors, 99 subsectors, and 311 industry groups \cite{wood2017automated}.

The FF system was initially conceptualized by academic researchers in finance as a means to investigate the industrial cost of capital \cite{fama1997industry}. FF achieves its purpose by reclassifying the existing SIC codes and grouping companies into 48 distinct industry sectors. Despite its prevalence in academic research concerning asset pricing, corporate finance, accounting, and economics, the FF system has not gained much popularity within the financial industry. In contrast, the Global Industry Classification Standard (GICS) was specifically designed by Standard \& Poor's (S\&P) and Morgan Stanley Capital International (MSCI) to meet the needs of financial professionals, such as investment managers and financial analysts. As shown in Table \ref{table:gics}, GICS employs an eight-digit code to classify companies, and its structure is hierarchical, encompassing 10 sectors subdivided into 24 industry groups, 64 industries, and 139 subindustries\footnote{A more detailed information about the GICS index is available at https://www.msci.com/our-solutions/indexes/gics}. The GICS scheme classifies companies based on their business activity, sources of revenue and earnings, as well as market perception concerning their primary lines of business \cite{chan2007industry}.

GICS has been shown to outperform other popular industry classification systems, such as SIC, NAICS, and FF, in various comparison experiments \cite{bhojraj2003s, kile2009using, hrazdil2012importance, boni2006analysts, chan2007industry}. This is due to its superior ability to capture industry homogeneity, leading to more accurate industry classifications. Additionally, the GICS index has been found to exhibit robust classification performance not only in settings with large and well-known companies (e.g. S\&P companies), but also when applied to smaller and less-followed companies \cite{hrazdil2013comparison}. This strong performance across a wide range of companies and industries makes the GICS index an ideal candidate for zero-shot learning contexts, as utilized in this paper.

Apart from the widely used mainstream classification schemes, there are several accessible alternatives that may not be as popular among institutional practitioners. The interested reader may refer to Bloomberg, Capital IQ (available on finance.yahoo.com), Hoovers \& First Research, Market Guide, MarketLine, Morningstar, and Thomson Reuters (available on msn.money.com)  \cite{lyocsa2009industry}. Other schemes include the Thomson Reuters Business Classification (TRBC), Industry Classification Benchmark (ICB), and International Standard Industrial Classification of All Economic Activities (ISIC) \cite{slavov2019company}.

\begin{table}[htbp!]
\caption{Modified sector names of the GICS taxonomy using TF-IDF preprocessing and removal of stop words with the purpose of increasing the F1 score in zero-shot classification.}
\begin{center}
\begin{tabular}{cc}
\hline
\multicolumn{2}{c}{\textbf{Sector names}}                                                                                    \\ \hline
Original GICS names    & Names after TF-IDF                                                                         \\ \hline
Energy                 & \begin{tabular}[c]{@{}c@{}}Oil, Natural Gas, Consumable Fuels\\ and Petroleum\end{tabular} \\
Materials              & \begin{tabular}[c]{@{}c@{}}Raw Materials, Mining, Minerals and\\ Metals (Gold, Silver and Copper)\end{tabular}    \\
Industrials            & Industrials and Transportation                                                             \\
Consumer Discretionary & \begin{tabular}[c]{@{}c@{}}Non-Essential Goods, Retail\\ and E-Commerce\end{tabular}                              \\
Consumer Staples       & \begin{tabular}[c]{@{}c@{}}Food, Beverages and\\ Household Products\end{tabular}           \\
Health Care            & Health Care                                                                                \\
Financials             & Banking and Lending                                                                        \\
Information Technology & Software, Technology and Systems                                                           \\
Communication Services & \begin{tabular}[c]{@{}c@{}}Communications, Telecommunications,\\ Networking, Media and Entertainment\end{tabular} \\
Utilities              & \begin{tabular}[c]{@{}c@{}}Utilities, Energy Distribution and\\ Renewable Energy\end{tabular}                     \\
Real Estate            & Real Estate Properties                                                                      \\ \hline
\end{tabular}
\end{center}
\label{table:sector_names}
\end{table}

\subsection{Issues with the current standards}

Despite the presence of diverse standards available for company classification and their wide use, the existing classification schemes suffer from several important limitations that deserve attention. Assigning companies to industries is currently performed manually and is vendor-specific \cite{lyocsa2009industry}. The process is time-consuming, subjective, and prone to inaccuracies, as existing classification schemes are often constructed and maintained by domain experts. These schemes can also become quickly outdated due to market developments and changes in products, technology, and business patterns. Therefore, relying solely on human-aided classification is not optimal. In fact, even with domain expertise and sufficient data, deciding which companies belong to an industry is not straightforward.

Inconsistencies in the classification process across different data vendors can pose issues in terms of accuracy and homogeneity, as highlighted by \cite{bhojraj2003s}. This is exemplified by the findings in \cite{kahle1996impact}, which analyzed manually assigned SIC codes for companies and revealed a significant discrepancy between two major data providers (Compustat and SRSP). Specifically, at the two-digit level, approximately 36\% of SIC classifications differed, while at the four-digit level, almost 80\% differed, as noted in \cite{lamby2018classifying}.

The timely update of classification schemes to reflect changing business and industry environments is another concern. Regardless of the underlying standard used, the classification scheme should incorporate current and frequently updated data. However, this poses a major challenge as adjusting and updating classification schemes to reflect changes in company structure and operations requires extensive human effort, which can be both time-consuming and expensive \cite{yang2016automatic}.

With multitude of standards available, it is not not trivial to select an appropriate industry classification scheme. The study in \cite{katselas2019know} highlights several issues in this regard. A major concern is that the different data sources used for classification exhibit mismatches even when applied within the same classification system. As a result, the same company can be classified in different partitions even under the same classification system if different data sources are used for the classification.

\section{Related work}
\label{sec:related_work}

Although some studies have been conducted in recent years, the literature on the application of NLP methods for industry classification remains limited overall.

In \cite{tagarev2019comparison}, the authors investigate the effectiveness of deep learning models on encyclopedic data from the English DBpedia Knowledge Base\footnote{https://wiki.dbpedia.org}. Specifically, the study evaluates the performance of two popular models, Glove and ULMfit, against two baseline models (one-hot unigram and one-hot bigram). The dataset used for the experiments includes 300,000 textual descriptions of companies from DBpedia. While the company descriptions are uniform in length, the dataset contains a significant variation in industry representation. The results indicate that the tested models perform well on the larger classes but exhibit a decline in performance on the smaller classes. Importantly, the study does not use a dataset that is considered a ``gold standard'', and DBPedia lacks an established industry taxonomy.

A relevant study, \cite{slavov2019company}, uses the same dataset and experimental setup as \cite{tagarev2019comparison}, but introduces BERT and XLNet models in addition to Glove and ULMfit. This study compares the four models with the same baseline (one-hot unigram and one-hot bigram) and finds that all algorithms perform acceptably on well-represented classes, but experience decreased overall performance on less represented classes. Although no algorithm stands out as the best for small classes, XLNet and BERT demonstrate more stable performance overall, thanks to their superior F1 scores. As noted in \cite{slavov2019company}, there are currently no benchmark datasets for industry classification. Previous studies rely on ``Industry Sector'' data comprising 6,000 company descriptions collected from the web and classified into 70 industry sectors, but they utilize algorithms that are considered outdated in modern big data applications; these algorithms include Naive Bayes, multinomial NB, Maximum Entropy classifier, Support Vector Machine, and k-Nearest Neighbors.

The authors of \cite{he2020exploring} investigate the usefulness of text-based industry classification using various word and document embedding techniques, in conjunction with different clustering algorithms. Their approach is applied to publicly traded companies in both the US and Chinese markets, and the results are compared against the GICS index. The study utilizes advanced embedding techniques such as BERT, but surprisingly, the results show that a simpler technique, Latent Semantic Indexing (LSI), combined with k-means clustering, outperforms BERT on two measures. This finding is remarkable because LSI, an extension of conventional techniques such as Bag of Words (BoW) and Term Frequency-Inverse Document Frequency (TF-IDF), is not commonly used in state-of-the-art (SOTA) text classification. As a result, this study sheds new light on the potential usefulness of LSI for text-based industry classification.

\begin{table}[htbp!]
\caption{Distribution of companies in the WRDS dataset across various GICS sectors.}
\begin{center}
\begin{tabular}{cc}
\hline
\multicolumn{2}{c}{\textbf{WRDS dataset}}    \\ \hline
GICS sector            & Number of companies \\ \hline
Energy                 & 2822                \\
Materials              & 3833                \\
Industrials            & 3934                \\
Consumer Discretionary & 4662                \\
Consumer Staples       & 1433                \\
Health Care            & 4565                \\
Financials             & 5363                \\
Information Technology & 5192                \\
Communication Services & 1285                \\
Utilities              & 740                 \\
Real Estate            & 509                 \\ \hline
\end{tabular}
\end{center}
\label{table:company_distribution_per_sector}
\end{table}

A method for fine-tuning a pre-trained BERT model is proposed in \cite{ito2020learning}, which is then evaluated on two datasets consisting of US and Japanese company data. The US dataset includes 2,462 annual reports from 2019 of companies listed on the US stock market (Form 10-K documents), while the Japanese dataset contains 3,016 annual reports from 2018 of companies listed on the Tokyo Stock Exchange. The paper's objective is to explore the extent to which companies with similar vector representations operate in comparable industries, as well as to evaluate how effectively companies can be classified within a given industry based solely on the industry name. The study compares BERT to two baseline models, namely, BoW representation and skip-gram Word2Vec embedding, and demonstrates BERT's superior performance. The findings confirm the effectiveness of the proposed approach and suggest the integration of additional sources of business data, such as Price Earnings Ratio (PER) and Price Book-value Ratio (PBR), to augment annual reports for the purpose of industry classification.

Other relevant references include \cite{yang2016automatic, wang2021enriching, wood2017automated, lamby2018classifying}, among others. \cite{yang2016automatic} introduced a new classification scheme called Business Text Industry Classification (BTIC), which was built on a dataset of Form 10-K documents from S\&P500 companies. Additionally, \cite{wang2021enriching} proposed a Knowledge Graph Enriched BERT (KGEB) model, which is capable of loading any pre-trained BERT model and fine-tuning it for classification. KGEB enriches word representations with additional knowledge through learning the graph structure in the underlying dataset. The model was tested on a dataset of publicly listed companies on the Chinese National Equities Exchange and Quotations (NEEQ). The authors of \cite{wood2017automated} employed a deep neural network based on a multilayer perceptron architecture with four fully-connected layers to predict the industries of novel companies. \cite{lamby2018classifying} evaluated 28 classifiers based on four underlying Word2Vec models with varying window sizes and different SVM kernels and logistic regression solvers. The Word2Vec models were trained on a corpus of articles from the Guardian newspaper with 600 million words. While the dataset is comprehensive, the paper does not present a comparison with benchmark schemes such as GICS. The use of Word2Vec is also somewhat outdated compared to SOTA deep learning models.

\section{Methodology and results}
\label{sec:methodology}

\subsection{Dataset}

We use the Wharton Research Data Services (WRDS) to create the dataset for this research. WRDS is a web-based data management system that provides researchers with access to a vast array of financial, economic, and marketing data from different sources, including Compustat, CRSP, IBES, and others. WRDS is a research platform that has been developed and maintained by the Wharton School at the University of Pennsylvania to support researchers in their data-driven research activities. The platform is used by academic researchers, corporate professionals, and financial analysts to retrieve, manage, and analyze large sets of data for their research projects. WRDS also provides a suite of tools for data cleaning, analysis, and visualization to help users get the most out of the data available on the platform. Using WRDS, we extract the Compustat dataset which contains financial and market data on publicly traded companies across the United States. The original dataset contains data for 44,033 companies, including their names, descriptions, and classification into sectors and industry groups as per the GICS taxonomy. We filter the dataset due to the absence of GICS sector assignment for some entries. After the filtering, the dataset consolidates a total of 34,338 entries (i.e. companies). The distribution of the number of companies across the GICS sectors is shown in Table \ref{table:company_distribution_per_sector}.

\subsection{Experiment}

We employ a zero-shot classification pipeline using the valhalla/distilbart-mnli-12-3 model. We chose the valhalla/distilbart-mnli-12-3 model for its popularity on Hugging Face, where it ranks as one of the top three models in terms of number of downloads. To assess its effectiveness, we conducted a comparative analysis with two other models, namely facebook/bart-large-mnli and joeddav/xlm-roberta-large-xnli. Our evaluation showed that the valhalla/distilbart-mnli-12-3 model performed slightly better than the other two models.

The valhalla/distilbart-mnli-12-3 model belongs to the class of transformer models which are a powerful type of neural network architecture that has been widely adopted in NLP \cite{wolf2020transformers}. Transformer models are capable of modeling long-range textual dependencies thereby effectively capturing relationships between distant words in a sentence \cite{devlin2018bert}. The transformer architecture includes an attention mechanism which allows the model to selectively focus on relevant parts of the input sequence. This enables the model to extract important relationships between words and better capture the meaning of the input text \cite{vaswani2017attention}.

Zero-shot classification refers to the ability of a model to classify inputs into multiple classes without requiring any training data \cite{pushp2017train}. Pre-trained transformer models have shown potential in zero-shot classification tasks as they have been trained on massive amounts of textual data, enabling them to classify inputs into classes even if they have never seen examples of those classes before. When needed, pre-trained models can be fine-tuned on specific zero-shot classification tasks with only a small amount of training data, allowing the models to adapt to the underlying task and improve the overall performance (e.g. accuracy) on that task.

\begin{figure}[htbp]
\centerline{\includegraphics[width=\columnwidth]{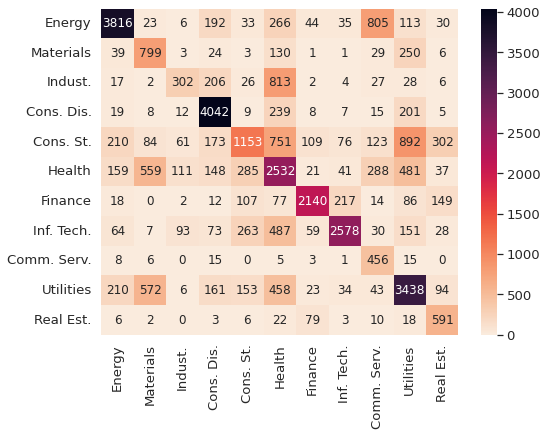}}
\caption{Confusion matrix of the zero-shot classification experiment using the valhalla/distilbart-mnli-12-3 model on the WRDS dataset.}
\label{fig:confusion_matrix}
\end{figure}

In our experiment, we utilized the valhalla/distilbart-mnli-12-3 model and adopted the zero-shot classification technique. Specifically, we fed the model with the company descriptions available in the WRDS dataset without performing any fine-tuning of the model. As we employed the zero-shot learning approach, we did not need to divide the dataset into train and test sets.

We obtained an F1 score of 0.56 using the original category names from the GICS taxonomy. We aimed to boost the F1 score by modifying the sector names with alternative labels that do not impede the model's classification performance. Specifically, we utilized TF-IDF vectorization to extract the top 30 most common words for each sector to obtain a more precise representation of the sector names, with the goal of enhancing the accuracy of the zero-shot classification model. This process involved pre-processing the company descriptions using the NLTK\footnote{https://www.nltk.org/} library to identify all verbs in the dataset and exclude them as stop words (since verbs are the most frequently occurring words). In addition to default stop words, we also excluded country names and certain abbreviations (e.g., ltd, llc) that occur frequently but are not relevant to the companies. The original and modified sector names are shown in Table \ref{table:sector_names}. This technique increases the F1 score to 0.64. However, the aforementioned change failed to result in any significant improvement in the sectors that had the lowest F1 scores, namely Real Estate, Consumer Staples, Consumer Discretionary, and Industrials.

The confusion matrix and classification report are presented on Figure \ref{fig:confusion_matrix} and in Table \ref{table:classification_report}. As can be seen, the weighted F1 score obtained across the dataset is 0.64. Highest individual F1 scores are obtained for Health Care and Oil \& Natural Gas with 0.84 and 0.81 F1 scores, followed by Banking \& Lending and Raw Minerals \& Mining with 0.77 and 0.75 F1 scores respectively. Lowest F1 scores are noticed for Food, Beverages and Household Products with 0.30 F1 score, Real Estate with 0.39 F1 score, and Industrials and Transportation with 0.39 F1 score.

\begin{table*}[htbp]
\caption{Classification report for the valhalla/distilbart-mnli-12-3 model on the WRDS dataset with original sector names.}
\begin{tabular}{rrrrr}
\multicolumn{1}{l}{}\phantom{Communications, Telecommunications, Networking, Media and Entertainment}                                                    & precision            & recall               & f1-score             & support              \\
\multicolumn{1}{l}{}                                                    & \multicolumn{1}{l}{} & \multicolumn{1}{l}{} & \multicolumn{1}{l}{} & \multicolumn{1}{l}{} \\
Financials                                 & 0.68 & 0.61  & 0.64 & 5363  \\
Communication Services                     & 0.32 & 0.63  & 0.42 & 1285  \\
Consumer Staples (Consumer Defensive)      & 0.20 & 0.01  & 0.02 & 1433  \\
Health Care                                & 0.83 & 0.84  & 0.83 & 4565  \\
Industrials                                & 0.42 & 0.20  & 0.27 & 3934  \\
Consumer Discretionary (Consumer Cyclical) & 0.41 & 0.46  & 0.43 & 4662  \\
Energy                                     & 0.56 & 0.91  & 0.69 & 2822  \\
Materials                                  & 0.54 & 0.65  & 0.59 & 3833  \\
Real Estate                                & 0.29 & 0.89  & 0.44 & 509   \\
Information Technology                     & 0.71 & 0.54  & 0.61 & 5192  \\
Utilities                                  & 0.44 & 0.25  & 0.32 & 740   \\
                                           &      &       &      &       \\
accuracy                                    &      &      & 0.56 & 34338 \\
macro avg                                  & 0.49 & 0.54  & 0.48 & 34338 \\
weighted avg                               & 0.57 & 0.56  & 0.55 & 34338 \\
                                           &      &       &      &      
\end{tabular}
\label{table:classification_report0}
\end{table*}

\begin{table*}[htbp]
\caption{Classification report for the valhalla/distilbart-mnli-12-3 model on the WRDS dataset with enhanced sector names.}
\begin{tabular}{rrrrr}
\multicolumn{1}{l}{}                                                    & precision            & recall               & f1-score             & support              \\
\multicolumn{1}{l}{}                                                    & \multicolumn{1}{l}{} & \multicolumn{1}{l}{} & \multicolumn{1}{l}{} & \multicolumn{1}{l}{} \\
Banking and Lending                                                     & 0.84                 & 0.71                 & 0.77                 & 5363                 \\
Communications, Telecommunications, Networking, Media and Entertainment & 0.39                 & 0.62                 & 0.48                 & 1285                 \\
Food, Beverages and Household Products                                  & 0.51                 & 0.21                 & 0.30                 & 1433                 \\
Health Care                                                             & 0.80                 & 0.89                 & 0.84                 & 4565                 \\
Industrials and Transportation                                          & 0.57                 & 0.29                 & 0.39                 & 3934                 \\
Non-Essential Goods, Retail and E-Commerce                              & 0.44                 & 0.54                 & 0.48                 & 4662                 \\
Oil, Natural Gas, Consumable Fuels and Petroleum                        & 0.86                 & 0.76                 & 0.81                 & 2822                 \\
Raw Materials, Mining, Minerals and Metals (Gold, Silver and Copper)    & 0.86                 & 0.67                 & 0.75                 & 3833                 \\
Real Estate Properties                                                  & 0.25                 & 0.90                 & 0.39                 & 509                  \\
Software, Technology and Systems                                        & 0.61                 & 0.66                 & 0.63                 & 5192                 \\
Utilities, Energy Distribution and Renewable Energy                     & 0.47                 & 0.80                 & 0.59                 & 740                  \\
                                                                        &                      &                      &                      &                      \\
accuracy                                                                &                      &                      & 0.64                 & 34338                \\
macro avg                                                               & 0.60                 & 0.64                 & 0.58                 & 34338                \\
weighted avg                                                            & 0.67                 & 0.64                 & 0.64                 & 34338               
\end{tabular}
\label{table:classification_report}
\end{table*}

\section{Conclusion}
\label{sec:conclusion}

We highlight the growing importance of natural language processing (NLP) in various business applications, including classification of companies. We propose an approach for company classification using NLP and zero-shot learning, which can streamline the process of categorizing companies and potentially reduce the time and resources required for traditional approaches such as the Global Industry Classification Standard (GICS). The proposed approach utilizes pre-trained transformer models to extract features from textual descriptions of companies and then applies zero-shot learning to classify them into relevant categories without the need for specific training data for each category. We evaluate the approach on the WRDS dataset and demonstrate its effectiveness in classifying companies. The results show that this method holds potential for automating the task of company classification, which can benefit various industries, including finance, marketing, and business intelligence, by providing a more efficient and cost-effective way of categorizing companies. It can also help in identifying emerging trends and patterns in the business world, which can be valuable for decision-making processes.

\bibliographystyle{IEEEtran}
\bibliography{references.bib}

\end{document}